\pdfoutput=1

\documentclass[11pt]{article}

\usepackage{ACL2023}

\usepackage{times}
\usepackage{latexsym}
\usepackage{microtype}
\usepackage{times}
\usepackage{latexsym}
\usepackage{amssymb}
\usepackage{amsfonts}
\usepackage{amsmath} 
\usepackage{amsthm} 
\usepackage{booktabs}
\usepackage{enumerate}
\usepackage{graphicx}
\usepackage{subfigure}
\usepackage{xspace}
\usepackage{float}
\usepackage{bbm}
\usepackage{bm}
\usepackage{multirow}
\usepackage{booktabs}
\usepackage{color}
\usepackage{framed}
\usepackage{stfloats}
\usepackage{iitem}
\usepackage{makecell}
\usepackage{float}
\restylefloat{table}
\usepackage{placeins}
\usepackage{afterpage}
\usepackage{pifont}
\usepackage[normalem]{ulem}
\useunder{\uline}{\ul}{}
\usepackage{url}

\newcommand{\ie}{\emph{i.e.,}\xspace}

\newcommand{\eg}{\emph{e.g.,}\xspace}

\newcommand{\ignore}[1]{}

\usepackage{xcolor}
\usepackage[T1]{fontenc}

\usepackage[utf8]{inputenc}

\usepackage{microtype}

\usepackage{inconsolata}

\title{Diffusion-NAT: Self-Prompting Discrete Diffusion for \\  Non-Autoregressive Text Generation}

\author{
\setcounter{footnote}{1}
	Kun Zhou\textsuperscript{\rm{1},\rm{3}},
	Yifan Li\textsuperscript{\rm{2}},
	Wayne Xin Zhao\textsuperscript{\rm{2},\rm{3}}\thanks{$^\dagger$ Corresponding author} \and
	\textbf{Ji-Rong Wen}\textsuperscript{\rm{2},\rm{3}} \\
	\textsuperscript{1}School of Information, Renmin University of China. \\
	\textsuperscript{2}Gaoling School of Artificial Intelligence, Renmin University of China \\
	\textsuperscript{3}Beijing Key Laboratory of Big Data Management and Analysis Methods\\
	\texttt{francis\_kun\_zhou@163.com}, \texttt{\{liyifan0925,batmanfly\}@gmail.com},\\
	\texttt{jrwen@ruc.edu.cn} \\
}

\begin{document}
\maketitle
\begin{abstract}
Recently, continuous diffusion models (CDM) have been introduced into non-autoregressive (NAR) text-to-text generation. 
However, the discrete nature of text increases the difficulty of CDM to generate coherent and fluent texts, and also causes the incompatibility problem between CDM and advanced NLP techniques, especially the popular pre-trained language models~(PLMs).
To solve it, we propose Diffusion-NAT, which introduces discrete diffusion models~(DDM) into NAR text-to-text generation and integrates BART to improve the performance.
By revising the decoding process of BART and the typical settings of DDM, we unify the inference process of BART and the denoising process of DDM into the same NAR masked tokens recovering task.
In this way, DDM can rely on BART to perform denoising, which can benefit from both the rich pre-learned knowledge of BART and the iterative refining paradigm of DDM.
Besides, we also propose the iterative self-prompting strategy to further improve the generation quality.
Experimental results on 7 datasets show that our approach can outperform competitive NAR methods, and even surpass autoregressive methods.
Our code and data will be publicly released.
\end{abstract}

\section{Introduction}

Text-to-text generation~\cite{DBLP:conf/nips/SutskeverVL14,DBLP:conf/nips/VaswaniSPUJGKP17} is an essential task in natural language processing, which aims to generate human-like texts satisfying the task demand. 
To efficiently generate high-quality texts, non-autoregressive~(NAR) models~\cite{DBLP:conf/iclr/Gu0XLS18,DBLP:conf/emnlp/LeeMC18} are widely explored for text-to-text generation by predicting all tokens in the target text simultaneously,  having a lower inference latency. 

\begin{table}[t]
	\small
	\centering
	\begin{tabular}{l|cccccc}
		\toprule
  Model & Type & PLMs & Cost & NAR & T2T \\
  \hline
  D3PM & Dis. & \ding{53} & Low & \ding{51} & \ding{53} \\
  Diffusion-LM & Con. & \ding{53} & Low & \ding{51} & \ding{53} \\
  SED & Con. & \ding{53} & Low & \ding{51} & \ding{53} \\
  SSD-LM & Con. & \ding{51} & High & \ding{51}  & \ding{53} \\
  DiffusionBERT & Dis. & \ding{51} & High & \ding{51} & \ding{53} \\
  LD4LG & Con. & \ding{51} & Low & \ding{53}  & \ding{53} \\
  \hline
  DiffuSeq & Con. & \ding{53} & Low & \ding{51} & \ding{51} \\
  SeqDiffuSeq & Con. & \ding{53} & Low & \ding{51} & \ding{51} \\
  GENIE & Con. & \ding{53} & High & \ding{51} & \ding{51} \\
  Difformer & Con. & \ding{53} & Low & \ding{51} & \ding{51} \\
  \hline
  Ours & Dis. & \ding{51} & Low & \ding{51}  & \ding{51} \\
		\bottomrule   
	\end{tabular}
	\caption{A comparison of existing diffusion methods for text generation. \textbf{Dis.} and \textbf{Con.} refer to discrete and continuous diffusion. \textbf{PLMs}, \textbf{Cost}, \textbf{NAR} and \textbf{T2T} denote using PLMs, Training Cost, Non-AutoRegressive model and Text-to-Text generation, respectively.}
	\label{tab:intro}
\end{table}

Despite the efficiency, the generation accuracy of NAR models generally underperform   autoregressive~(AR) models with the token-by-token generation, since parallel token prediction cannot effectively capture the dependency among the tokens. 
To enhance the generation quality, a variety of techniques have been proposed for  NAR models, with either improved architectures~\cite{DBLP:conf/acl/QianZBWQ0YL20} or training methods~\cite{DBLP:conf/icml/QiG0YCLTLCZ0D21}. More recently, inspired by the success of diffusion models in computer vision~\cite{DBLP:conf/nips/HoJA20,DBLP:conf/nips/DhariwalN21}, they have been introduced to improve NAR models for text-to-text generation~\cite{li2023diffusion,DBLP:journals/corr/abs-2212-11685,DBLP:journals/corr/abs-2210-08933}. 
As shown in Table~\ref{tab:intro}, these studies typically adopt the continuous diffusion method on the latent space of token embeddings in the NAR manner, and iteratively refine all the target token embeddings via a parameterized denoising process.

However, these attempts  are highly limited by the discrete nature of text, and thus it is necessary to incorporate special  strategies to adapt  continuous diffusion models for text generation.    
Typically, they rely on an additional rounding step~\cite{DBLP:journals/corr/abs-2205-14217} to map the generated embeddings into tokens, and add corresponding loss during training.
However, the added step and training loss would burden the diffusion models, causing them hungry for more training steps and data to capture the mapping relation between input and output.
Although large-scale pre-trained language models~(PLMs)~\cite{DBLP:conf/naacl/DevlinCLT19,DBLP:conf/acl/LewisLGGMLSZ20} seem to be a promising solution to alleviate this hunger problem, 
due to the large model discrepancy, it is difficult to use existing PLMs for improving the text generation models when integrating with continuous diffusion models,
even leading to performance degradation~\cite{DBLP:journals/corr/abs-2205-14217}.

\ignore{
For continuous diffusion models, the discrete nature of the text has become an awkward problem.
Previous works~\cite{DBLP:journals/corr/abs-2205-14217} rely on the rounding step to map the generated embeddings into tokens, and adds the corresponding objective for training.
Such a way burdens the diffusion models, making them hungry for more training data and steps to characterize the relations between input and output and the dependency among contextual tokens.
As a promising solution, large-scale pre-trained language models~(PLMs)~\cite{DBLP:conf/naacl/DevlinCLT19,DBLP:conf/acl/LewisLGGMLSZ20} might be able to serve as a good starting point for initializing diffusion models.
However, due to the inconsistent learning objectives and model architectures, these continuous diffusion models can not efficiently and effectively benefit from the existing PLMs, even leading to performance degradation~\cite{DBLP:journals/corr/abs-2212-09412}.
}

To address these issues, we aim to develop a more effective approach to integrating diffusion models and PLMs for NAR text-to-text generation. 
Instead of using continuous diffusion, we utilize discrete diffusion~\cite{DBLP:conf/nips/AustinJHTB21,DBLP:conf/cvpr/GuCBWZCYG22} for text generation, which performs denoising on discrete states (\eg vocabulary) to recover the original tokens. 
It is more suitable for modeling discrete text data, making it potentially feasible to develop more \emph{unified and compatible} solutions to integrate diffusion models and well-trained PLMs for improving NAR text generation models.    
However, both discrete diffusion models and PLMs neither naturally fit with each other nor the NAR text-to-text generation manner, making it hard to directly combine them for improving the NAR generation quality. 
In this paper, we propose \textbf{Diffusion-NAT}, a self-prompting discrete diffusion model using PLMs for NAR text-to-text generation.
The core contribution lies in that we unify the \emph{inference process} of PLMs and \emph{denoising process} of discrete diffusion models into the same masked token recovering task in the NAR manner.
In this way, PLMs can play the role of the parameterized denoiser in discrete diffusion models, hence we can combine the merits of both diffusion models (\emph{using iterative refining generation}) and PLMs (\emph{with rich semantic knowledge}) for improving NAR text generation. 
Specifically, we select the Seq2Seq PLM, BART~\cite{DBLP:conf/acl/LewisLGGMLSZ20} as our backbone by revising its decoding process into the NAR masked tokens recovering task, and adjust the typical discrete diffusion method to better fit the PLM by adding mask tokens as  noise, revising the learning objective and removing the time step embeddings.
Further, as our approach performs the denoising process fully based on the PLM, we devise an iterative self-prompting strategy to guide the PLM performing multi-turn deliberation and refinement on the intermediate generated results, to enhance the quality of the final output.



To demonstrate the effectiveness of our approach, we conduct extensive experiments on seven text-to-text generation datasets. 
Experimental results show that our approach can outperform competitive NAR text generation methods, \eg improving the best NAR models by +2.48 BLEU-2 on PersonaChat, +4.33 Distinct-2 on DailyDialog.
Our approach even surpasses state-of-the-art autoregressive PLMs, \eg Our (27.90) VS. BART (23.54) Overall metrics on PersonaChat, and Our (44.2) VS. BART (38.3) ROUGE-L on MSNews.


\section{Related Work}
\paragraph{Non-Autoregressive Text Generation.}
Compared with autoregressive~(AR) methods~\cite{DBLP:conf/acl/LewisLGGMLSZ20} that need to predict the target text in a token-by-token manner, Non-autoregressive~(NAR) methods can generate all tokens in parallel, which can greatly reduce the inference latency~\cite{DBLP:conf/iclr/Gu0XLS18,DBLP:conf/emnlp/GhazvininejadLL19}.
However, in this way, NAT methods can not fully capture the dependency relations among tokens during decoding, leading to the sacrifice of the accuracy.
To address it, existing works adopt several training and inference strategies to improve the performance of NAR methods, \eg knowledge distillation~\cite{DBLP:conf/iclr/ZhouGN20}, glancing sampling~\cite{DBLP:conf/acl/QianZBWQ0YL20}, iterative decoding~\cite{DBLP:conf/emnlp/GengF021} and large-scale pre-training~\cite{DBLP:conf/icml/QiG0YCLTLCZ0D21, DBLP:journals/corr/abs-2210-13304}.
In this work, we introduce the discrete diffusion model into NAR text generation, narrowing the performance gap with AR methods.

\paragraph{PLMs for Text Generation.}
Pre-trained language models~(PLMs) have shown remarkable performance in generating human-like texts~\cite{DBLP:journals/corr/abs-2105-10311}.
After pre-training, most existing PLMs~\cite{DBLP:journals/jmlr/RaffelSRLNMZLL20} are fine-tuned following the AR paradigm for text generation.
In this way, they either reformulate generation tasks into the language model format (\eg GPT~\cite{radford2019language}), or leverage the sequence-to-sequence manner to generate the text using an autoregressive decoder (\eg BART~\cite{DBLP:conf/acl/LewisLGGMLSZ20}).
However, as these PLMs only focus on fine-tuning under the AR paradigm, they can not be directly used for NAR text generation.
Recently, BANG~\cite{DBLP:conf/icml/QiG0YCLTLCZ0D21} and ELMER~\cite{DBLP:journals/corr/abs-2210-13304} rely on large-scale pre-training for improving the NAR text generation.
Considering the pre-training cost, we aim to efficiently adapt BART into an effective NAR model with diffusion models.

\paragraph{Diffusion Models for Text Generation.}
Diffusion models~(DM)~\cite{DBLP:conf/nips/HoJA20, DBLP:conf/iclr/0011SKKEP21} are a class of latent variable models that can progressively denoise a random Gaussian noise into a data example.
Existing DMs can be roughly categorized into continuous diffusion models~\cite{DBLP:conf/nips/HoJA20} and discrete diffusion models~\cite{DBLP:conf/nips/AustinJHTB21}, which perform diffusion on continuous signals and discrete states, respectively.
Recently, DMs have been utilized for text generation and have demonstrated superiority in controllable text generation tasks~\cite{li2023diffusion,DBLP:journals/corr/abs-2205-14217}.
For text-to-text generation tasks, 
existing works generally follow the continuous diffusion paradigm, and improve the performance by refining the model architecture~\cite{DBLP:journals/corr/abs-2212-10325}, adding regularization~\cite{DBLP:journals/corr/abs-2212-09412} and large-scale pre-training~\cite{DBLP:journals/corr/abs-2212-11685}.
In this work, we aim to introduce discrete diffusion model into text-to-text generation tasks, and utilize a PLM to improve it.

\section{Preliminary}

\paragraph{Problem Statement.}
\label{sec-problem}
This work focuses on text-to-text generation tasks using non-autoregressive (NAR) models.
Generally, text-to-text generation tasks~\cite{DBLP:conf/nips/SutskeverVL14,DBLP:conf/nips/VaswaniSPUJGKP17} (\eg dialog and summarization) can be formulated as modeling the conditional probability $P(Y|C)$, where $C=\{c_{1}, c_{2}, \cdots, c_{m}\}$ and $Y=\{y_{1}, y_{2}, \cdots, y_{n}\}$ denote the input text and output text respectively, both consisting of a sequence of tokens from a vocabulary $\mathcal{V}$. 

Different from AR models with the left-to-right token-by-token generation manner, NAR models~\cite{DBLP:conf/iclr/Gu0XLS18,DBLP:conf/emnlp/LeeMC18} predict all tokens of the output text $Y$ simultaneously, where each token $y_i$ is predicted only based on the input text $C$. Therefore, the conditional probability can be factorized as
\begin{equation}
    P(Y|C)=\prod_{i=1}^{n}P(y_{i}|C),
    \label{eq-CE}
\end{equation}


\paragraph{Diffusion Models.}
Diffusion models~(DM)~\cite{DBLP:conf/nips/HoJA20,DBLP:conf/iclr/0011SKKEP21} sample an example from a data distribution $p(x)$ by gradually denoising a random noise.
Typically, starting from a noise $x_{T}$, the denoising process (also so-called reverse process) can be regarded as a Markov process, where the noises at $T-1, T-2, \cdots, 0$ steps are progressively predicted and removed to obtain the latent variables $x_{T-1}, x_{T-2}, \cdots$, until reaching the final sample $x_0$. 
Conversely, given the sample $x_0$, we can generate $x_{1}, x_{2}, \cdots, x_{T}$ as a Markov chain, denoted as the \emph{forward process}: 
\begin{equation}
    q(x_{t}|x_{t-1})=\mathcal{N}(\sqrt{1-\beta_{t}}x_{t-1}, \beta_{t}\emph{\textbf{I}}),
\end{equation}
where $\beta_t \in (0,1)$ is the pre-defined scaling of noise variance at the $t$-th step. 
Given the above forward process as prior, DMs are trained to reverse it following the denoising process for recovering $x_0$, where each step is parameterized as:
\begin{equation}
    p(x_{t-1}|x_{t})=\mathcal{N}(\mu_{\theta}(x_t, t), \Sigma_{\theta}(x_t, t)),
\end{equation}
where $\mu_{\theta}(\cdot)$ and $\Sigma_{\theta}(\cdot)$ can be implemented by a U-Net~\cite{DBLP:conf/miccai/RonnebergerFB15} or Transformer~\cite{DBLP:conf/nips/VaswaniSPUJGKP17}, and time step embeddings are adopted to represent $t$.

\paragraph{Discrete Diffusion Models.}
Discrete diffusion models~\cite{DBLP:conf/nips/AustinJHTB21,DBLP:conf/cvpr/GuCBWZCYG22} perform the forward and denoising processes in discrete random variables with $K$ categories, where $K=|\mathcal{V}|$ for text data.
For a sentence, $x_0$ is the vector consisting of the indexes of its contained tokens, and the forward process of adding noise is 
\begin{equation}
    q(x_{t}|x_{t-1})=v^{\top}(x_{t})\mathbf{Q}_{t}v(x_{t-1}),
\end{equation}
where $v(x_t)$ maps each token index from $x_t$ into $K$-dimension one-hot vector, $\mathbf{Q}_{t}$ is the probability transition matrix and $[\mathbf{Q}_{t}]_{i,j}$ denotes the probability of the token $i$ to be replaced by the token $j$.
In this way, according to Bayes' theorem, the denoising process $q(x_{t-1}|x_{t},x_0)$ can be deduced as:
\begin{equation}
\frac{(v^{\top}(x_{t})\mathbf{Q}_{t}v(x_{t-1}))(v^{\top}(x_{t-1})\bar{\mathbf{Q}}_{t-1}v(x_{0}))}{v^{\top}(x_{t})\bar{\mathbf{Q}}_{t}v(x_{0})}
\end{equation}
where $\bar{\mathbf{Q}}_{t}=\mathbf{Q}_{1}\mathbf{Q}_{2}\cdots\mathbf{Q}_{t}$.
Based on the above prior, we can use a parameterized model $p_{\theta}(x_{t-1}|x_{t},t)$ to learn the denoising process.
\begin{figure*}[t]
\centering
\includegraphics[width=0.95\textwidth]{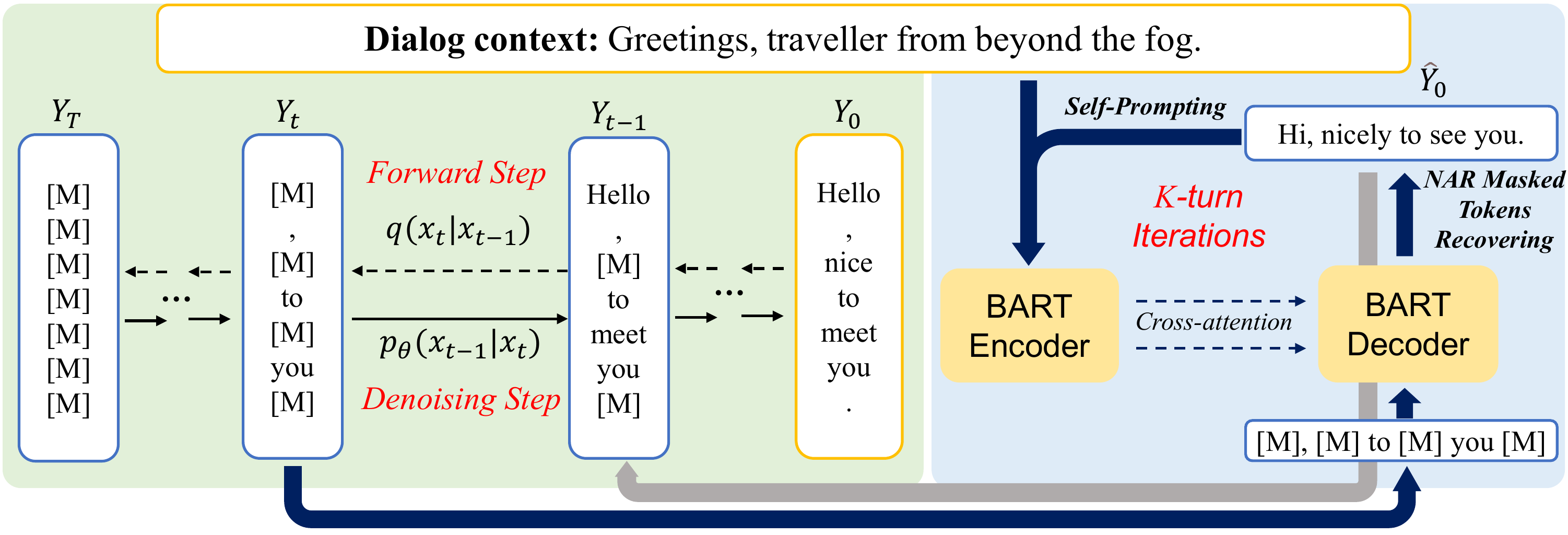}
\caption{The overview of our Diffusion-NAT. We show an example that generates a response in the $t$-th step using $K$-turn self-prompting. 
The given dialog context and the $K$-turn prompt (\ie estimated $\hat{Y}_0$) are fed into BART encoder, and the response in the $t$-th $Y_t$ is fed into BART decoder for estimating the original tokens.}
\label{fig:framework}
\end{figure*}

\section{Approach}
In this section, we introduce Diffusion-NAT, an effective approach to integrating the discrete diffusion model and the  Seq2Seq PLM BART, for improving NAR text-to-text generation. The overview of our approach is shown in Figure~\ref{fig:framework}.

\subsection{Overview}

Since discrete diffusion models~(DDM) and BART adopt different ways for training (\ie noise prediction and masked text infilling respectively), it is hard to directly integrate both for NAR text-to-text generation. Our solution is to regard the mask token $\texttt{[MASK]}$ of BART as the \emph{noise} in DDM, and incorporate an absorbing state $\texttt{[MASK]}$ into the Markov transition matrices. In this way, the forward process of DDM  gradually replaces all the tokens by $\texttt{[MASK]}$, and the denoising process can be reformulated as a \emph{NAR Masked Tokens Recovering~(NMTR)} task: 
\begin{equation}
    f_{\text{NMTR}}({[\texttt{M}], \cdots, [\texttt{M}]})=\{y_1, \cdots, y_n\},
    \label{eq-general-task}
\end{equation}
where $[\texttt{M}]$ denotes the $\texttt{[MASK]}$ token of BART. 

To apply this framework for NAR text generation, we further make adaptations for BART and DDM.
For BART, its pre-training task of masked text infilling is similar to the above objective expect that it is in a NAR manner, and thus we revise the decoding process of BART to support the NAR inference in Section~\ref{sec-BART}.
For DDM, we learn to predict the original tokens instead of noise and remove the time step embeddings in Section~\ref{sec-Diffusion}, for better adaptation to BART. 
In this way, we can unify the inference process of BART and the denoising process of discrete diffusion models with the same formulation of \emph{NAR masked tokens recovering}.  


With this unified formulation, DDM can fully rely on BART to conduct the denoising process, with no need of additional parameters or specific training. In this way, the generated results based on BART can be iteratively refined via the denoising process, leading to improved generation text. Since BART is employed as the backbone of our approach, we can naturally leverage advanced techniques of PLMs to improve the diffusion process, \eg prompt learning~\cite{DBLP:journals/corr/abs-2107-13586}. Thus, we propose the iterative self-prompting strategy to perform multi-turn deliberation and refinement on the intermediate generated results in Section~\ref{sec-Prompt}, further enhancing the quality of the output.

\subsection{Adapting BART for NAR Generation}
\label{sec-BART}
\ignore{BART~\cite{radford2019language,DBLP:conf/acl/LewisLGGMLSZ20} adopts a token-by-token autoregressive way for decoding. 
Such a way is inconsistent with the NAR text generation task and would be unaffordable if combining with the iterative denoising process of diffusion models.
Therefore, we adjust it as the NAR masked tokens recovering process, which is similar to the pre-training task of BART.}

Since BART utilizes a token-by-token autoregressive mechanism for decoding, this part discusses how to  revise its decoding process to fit the NAR generation framework. 


\paragraph{BART.}
BART~\cite{DBLP:conf/acl/LewisLGGMLSZ20} is a Seq2Seq PLM that has been widely used on various text-to-text generation tasks.
It adopts the encoder-decoder Transformer architecture.
Given the input text $C$, the encoder produces its representation vectors $\mathbf{E}$, and the decoder performs cross-attention with $\mathbf{E}$ to inject the condition from the input text.
During pre-training, the masked text infilling task is mainly adopted to learn the model parameters on a large-scale corpus, aiming to  recover the masked span from the input text.
During inference, using a special start token as the initial input of the decoder, the output text will be generated token by token.

\paragraph{Revised NAR Decoding Process.}
In the denoising process of our approach,  BART is employed  to recover the masked tokens 
from the noised target text at each time step.
Thus, we revise the decoding process of BART into the NAR manner that can recover all masked tokens simultaneously.
Concretely, at the $t$-step, given the condition text $C$ and the noised target text $Y_t$ containing $\texttt{[MASK]}$ tokens, we feed them into the encoder and decoder of BART respectively, and simultaneously recover all the $\texttt{[MASK]}$ tokens into the target tokens as:
\begin{equation}
    \text{BART}(\{y^{(t)}_{1} \cdots [\texttt{M}]\}, C)=\{y^{(t-1)}_{1} \cdots y^{(t-1)}_{n}\},
    \label{eq-nar-bart}
\end{equation}
where $y^{(t)}_{1}$ is the token of the first position at the $t$-th step.
In this way, the decoding process follows the unified formulation in Eq.~\ref{eq-general-task}.  Thus, we can employ BART in the denoising  process by leverage its pre-learned knowledge and generation capacity.  


\subsection{Adapting DDM for NAR Generation}
\label{sec-Diffusion}
In this part, we discuss how to adapt discrete diffusion model~(DDM) to NAR masked tokens recovering for text generation. 


\paragraph{Markov Transition Matrices with $\texttt{[MASK]}$.} As introduced in Section~\ref{sec-problem}, discrete diffusion models rely on the probability transition matrix $\mathbf{Q}_{t}$ to perform the forward and denoising processes over the state space.  
 To align DDM with the NAR decoding process of BART~(Section~\ref{sec-BART}), we incorporate the $\texttt{[MASK]}$ token as the absorbing state of the Markov transition matrices. 
Concretely, at the $t$-th step of the forward process, if token $i$ is not the $\texttt{[MASK]}$ token, it has the probabilities of $\alpha_t$ and $\gamma_t$ being unchanged and replaced by the $\texttt{[MASK]}$ token respectively, leaving the probability of $\beta_t=1-\alpha_t-\gamma_t$ transiting to other tokens in $\mathcal{V}$ as:
\begin{equation}
    [\mathbf{Q}_{t}]_{i,j}=\left\{
\begin{aligned}
&\alpha_{t},& &\text{if} \   j=i,& \\
&\gamma_{t},& &\text{if} \   j=[\texttt{M}],& \\
&1-\alpha_t-\gamma_t,& &\text{otherwise},&
\end{aligned}
\right.
\end{equation}
where $\alpha_t$ and $\gamma_t$ are determined by the pre-defined noise schedule, \eg cosine schedule~\cite{DBLP:conf/icml/NicholD21}.
While, if token $i$ is the $\texttt{[MASK]}$ token, it will be unchanged. 
Based on such forward process, all tokens in the output text would become $\texttt{[MASK]}$ after a sufficient number of steps, corresponding to the all-$\texttt{[MASK]}$ input in Eq.~\ref{eq-general-task}.
In the denoising process, we adopt BART to gradually recover the all-$\texttt{[MASK]}$ sequence into output text in the NAR manner, where each denoising step is equivalent to the decoding of BART in Section~\ref{sec-BART}. 

\paragraph{Training with NAR Masked Tokens Recovering.}
During training, existing diffusion models mostly learn to predict the noise in the current time step.
However, such training objective is not consistent with PLMs.
Inspired by existing works~\cite{DBLP:journals/corr/abs-2205-14217,DBLP:journals/corr/abs-2210-08933}, we aim to predict all the original tokens $Y_0=\{y^{(0)}_1, \cdots, y^{(0)}_n\}$ using BART in the NAR manner at each time step as:
\begin{equation}
    \text{BART}(\{y^{(t)}_{1} \cdots [\texttt{M}]\}, C)=\{y^{(0)}_{1} \cdots y^{(0)}_{n}\}.
    \label{eq-nar-x0}
\end{equation}
As $Y_t$ usually contains several $\texttt{[MASK]}$ tokens, the above process can be regarded as recovering all the masked tokens into the original ones, which is actually similar to the pre-training objective of BART.
In this way, the training objective is formulated as:
\begin{equation}
    \mathcal{L}_{Y}=
    -\sum_{i=1}^{n}\log p_{\theta}(y^{(0)}_i|Y_t, C)
    \label{eq-mlm}
\end{equation}
where $Y_t$ denotes the intermediate recovered text in the $t$-th step.
During inference, given $Y_t$, our model first estimates $\hat{Y}_0$, and then adds the $(t-1)$-step noise into it for producing $Y_{t-1}$.
The above process will  iterate for multiple steps, until the final results of $Y_0$  are obtained.

\paragraph{Removing Time Step Embeddings.}
As another difference in architecture,  diffusion models typically incorporate time step embeddings to represent the time information~\cite{DBLP:conf/nips/HoJA20,DBLP:conf/iclr/SongME21}, while BART has never set up corresponding time embeddings.  
To reduce such discrepancy, we directly remove the time step embeddings from our diffusion process, so as to adapt DDM to reusing the whole architecture and all pre-trained parameters of BART.  Actually, as the discrete diffusion process is to progressively recover the all-$\texttt{[MASK]}$ sequence, the PLM can directly acquire the time information by counting the number of $\texttt{[MASK]}$ tokens. 
Further, by removing the time embeddings, our diffusion approach can better integrate with other improvement techniques, \eg DDIM method~\cite{DBLP:conf/iclr/SongME21} with the non-Markov process for fast inference.


\ignore{Actually, as the discrete diffusion process is to progressively recover the all-$\texttt{[MASK]}$ sequence, the PLM can directly acquire the time information by counting the number of $\texttt{[MASK]}$ tokens.
Therefore, we remove the time step embeddings in our approach.
In this way, we can directly reuse the architecture and pre-trained parameters of the PLM.
Furthermore, as such a way allows PLMs to ignore the current time step during performing iterative denoising, it also better suits the commonly-used DDIM method~\cite{DBLP:conf/iclr/SongME21} that incorporates the non-Markov process for the fast inference of diffusion models.
}

\subsection{Iterative Self-Prompting}
\label{sec-Prompt}
In a typical denoising process, the denoising network relies on the condition $C$ and $Y_t$ to estimate $\hat{Y}_0$.
However, at early steps, $\texttt{[MASK]}$ tokens generally occupy the majority of $Y_t$, causing the estimation much difficult.
To reduce the inference difficulty at an early stage, we propose the iterative self-prompting strategy that endows our model with deliberation capacity via prefixed prompts. 


\paragraph{Training with Self-Prompting.}
Inspired by the self-conditioning strategy~\cite{DBLP:journals/corr/abs-2208-04202}, our self-prompting strategy focuses on improving the quality of $\hat{Y}_0$ through multi-round checking and revision.
Concretely, given $Y_t$ and $C$, we first utilize the PLM to produce the estimated $\hat{Y}_0$.
Then, as $\hat{Y}_0$ and $C$ are two sequences of tokens, we regard $\hat{Y}_0$ as the prompt of the PLM and prefix it with $C$ to compose the new input condition $C'=[\hat{Y}_0;C]$.
Next, the new condition $C'$ and $Y_t$ are further fed into the encoder and decoder of the PLM respectively, where cross-attention in the decoder is employed to generate $\hat{Y}_0$ by considering the previous estimation.    
During training, with a certain probability (\eg 50\%), we do not use the self-prompting strategy and only optimize the model parameter using Eq.~\ref{eq-mlm}.
When integrated with this strategy, we first produce $\hat{Y}_0$ and then construct $C'$ for self-prompting, where the training objective becomes:
\begin{equation}
    \mathcal{L}_{Y}=-\sum_{i=1}^{n}\log p_{\theta}(y^{(0)}_i|Y_t, \hat{Y}_0, C). 
\end{equation}

\paragraph{Inference with Iterative Self-Prompting.}
To obtain a well-estimated $\hat{Y}_0$, we repeat the following self-prompting process for $K$ times: we first estimate the original tokens $\hat{Y}_0=\{\hat{y}^{(0)}_{1}, \cdots, \hat{y}^{(0)}_{n} \}$ based on the constructed new condition $C'$ and then utilize it to replace the original prompt within $C'$.
Each of the iterative process can be denoted as:
\begin{equation}
    \text{BART}\big(\{y^{(t)}_{1} \cdots \}, \{\hat{y}^{(0)}_{1} \cdots \}, C\big)=\{y^{(0)}_{1} \cdots \}. \nonumber
    \label{eq-nar-x0}
\end{equation}
In this way, by setting proper hyper-parameter $K$, we can balance the accuracy of the estimated $\hat{Y}_0$ and the time cost during inference.
Note that such manner also supports the explicit control in the intermediate prompts for guiding the generation, \eg correcting grammar errors in $\hat{Y}_0$.
We leave it as our future work.

\section{Experiments}
\subsection{Experimental Settings}
In this part, we present the datasets and evaluation metrics used in our experiments.
More details about the datasets, baselines, and implementations are shown in Appendix~\ref{app-dataset}, \ref{app-baselin} and \ref{app-details}, respectively.

\begin{table}[t]
    \small
    \centering
    \begin{tabular}{c|l|rrr}
        \toprule
             \textbf{Task}& \textbf{Datasets}& \#Train & \#Valid & \#Test   \\
        \hline
             \multirow{2}{*}{\textbf{Dialog}}&\textbf{DailyDialog}& 76,052 & 7,069 & 6,740 \\
             &\textbf{PersonaChat}& 122,499 & 14,602 & 14,056 \\
             \hline
             \multirow{2}{*}{\textbf{Sum.}}&\textbf{XSUM}& 204,045 & 11,332 & 11,334 \\
             &\textbf{MSNews}& 136,082 & 7,496 & 7,562 \\
             \hline
             \multirow{2}{*}{\textbf{QG}}&\textbf{MSQG}& 198,058 & 11,008 & 11,022 \\
             &\textbf{SQUAD v1.1}& 75,722 & 10,570 & 11,877 \\
             \hline
             {\textbf{CQA}}&\textbf{CoQA}& 108,647 & 3,935 & 4,048 \\
        \bottomrule
    \end{tabular}
    \caption{Statistics of the datasets, where \textbf{Dialog}, \textbf{Sum.}, \textbf{QG} and \textbf{CQA} denote Dialog Generation, Text Summrization, Question Generation and Conversational Question Answering, respectively.}
    \label{tab:statistics}
\end{table}

\paragraph{Datasets.}
We conduct experiments on 7 datasets, \ie DailyDialog, PersonaChat, XSUM, MSNews, MSQG, SQuAD v1.1 and CoQA, corresponding to four representative text generation tasks. Their statistics are shown in Table~\ref{tab:statistics}. 

\begin{table*}[t]
	\small
	\centering
	\begin{tabular}{l l  cccccccccc}
		\toprule
		\multirow{2.5}{*}{\textbf{Type}} & \multirow{2.5}{*}{\textbf{Models}} & \multicolumn{5}{c}{\textbf{PersonaChat}} & \multicolumn{5}{c}{\textbf{DailyDialog}} \\ 
		\cmidrule{3-12}
		&  & B-1$\uparrow$  & B-2$\uparrow$ & D-1$\uparrow$ & D-2$\uparrow$ & Overall$\uparrow$ & B-1$\uparrow$  & B-2$\uparrow$ & D-1$\uparrow$ & D-2$\uparrow$ & Overall$\uparrow$ \\ 
		\midrule
		\multirow{4}{*}{\textbf{AR}} & \textbf{Transformer} & 41.56 & 32.95 & 0.30 & 0.80 & 18.90 & 45.95 & 40.60&  0.91 & 4.68 & 23.04\\
		& \textbf{MASS} & 41.06 & 35.75 & 1.40 & 6.90 & 21.28 & 51.77 & 45.09& 3.99 & 23.38 &31.06 \\
		& \textbf{ProphetNet} & 46.00 & 38.40 & 1.30 & 7.30 & 23.25 & - & - & - & - & -\\ 
		& \textbf{BART} & 47.60 & 39.36 & 1.10 & 6.10 & 23.54 & 56.18 &49.59& 5.04 & 27.72 &34.63 \\ 
		\cmidrule{1-12}
		\multirow{5}{*}{\textbf{Semi-NAR}} & \textbf{InsT} & 12.63 & 9.43 & 0.10 & 0.30 & 5.62 & - & - & - & - & - \\
		& \textbf{iNAT} & 41.17 & 32.13 & 0.10 & 1.10 & 18.63 & - & - & - & - & - \\
  & \textbf{LevT} & 24.89 & 18.94 & 0.10 & 0.60 & 11.13 & - & - & - & - & - \\
		& \textbf{CMLM} & \underline{44.38} & \underline{35.18} & 0.10 & 0.80 & 20.12 & - & - & - & - & -  \\
		& \textbf{BANG} & 39.82 & 30.72 & 1.90 & 14.20  & \underline{21.66}  & 41.47 &35.71 &1.76 & 13.98 & 23.23 \\
		\cmidrule{1-12}
		\multirow{6.5}{*}{\textbf{NAR}} & \textbf{NAT} & 31.53 & 24.17 & 0.10 & 0.80 & 14.15 & - & - & - & - & - \\
		& \textbf{iNAT} & 30.56 & 23.38 & 0.10 & 0.70  & 13.69 & - & - & - & - & - \\
		& \textbf{CMLM} & 31.44 & 24.06 & 0.10 & 0.60  & 14.05  & - & - & - & - & - \\
		& \textbf{LevT} & 26.92 & 20.47 & 0.00 & 0.40  & 11.95  & - & - & - & - & -\\
		& \textbf{BANG} & 31.11 & 23.90 & 2.50 & 22.70 & 20.05  & 35.50 & 30.15 & 1.90 & 15.13 & 20.67\\
		& \textbf{ELMER} & 31.45 & 23.99 & \textbf{3.66} & \underline{24.96} & 21.02 & 68.32 & 61.14 &5.30 &35.64 & 42.60 \\
		\cmidrule{1-12}
  \textbf{Diffusion} & \textbf{Ours} & \textbf{44.55} & \textbf{37.66} & \underline{3.19} & \textbf{26.20} & \textbf{27.90} & \textbf{68.79} & \textbf{62.68} & \textbf{6.67} & \textbf{39.97} & \textbf{44.53} \\
		\bottomrule   
	\end{tabular}
	\caption{The comparison between our approach and baselines on two dialog generation tasks. B-1/2 and D-1/2 denote BLEU-1/2 and Distinct-1/2. \textbf{Bold} and \underline{underline} fonts denote the best and second best methods within NAR and Semi-NAR models, respectively. The baseline results on PersonaChat are collected from \cite{DBLP:journals/corr/abs-2210-13304}.}
	\label{tab:personachat}
\end{table*}

\paragraph{Evaluation Metrics.}
Following existing works \cite{DBLP:journals/corr/abs-2210-13304,DBLP:conf/icml/QiG0YCLTLCZ0D21}, we employ corresponding metrics to evaluate model performances on different tasks. For dialog generation, we adopt BLEU-1/2~\cite{DBLP:conf/acl/PapineniRWZ02} to measure the coherence between the generated and real responses based on the co-occurrence ratio of $n$-grams, and Distinct-1/2~\cite{DBLP:conf/naacl/LiGBGD16} to measure the $n$-gram diversity of the generated texts. 
For text summarization, we utilize ROUGE-1/2/L~\cite{lin2004rouge} to compute the overlapping ratio of $n$-grams between the generated and ground-truth summarizations for estimating the quality.
For question generation, we use ROUGE-L, BLEU-4 and METEOR~\cite{DBLP:conf/acl/BanerjeeL05} to assess the generation consistency. 
For conversational question answering, we adopt F1-Score~\cite{DBLP:conf/emnlp/RajpurkarZLL16} to measure the prediction accuracy.



\begin{table*}[t]
	\small
	\centering
	\begin{tabular}{l l  r r r r r r}
		\toprule
		\multirow{2.5}{*}{\textbf{Type}} & \multirow{2.5}{*}{\textbf{Models}} & \multicolumn{3}{c}{\textbf{XSUM}} & \multicolumn{3}{c}{\textbf{SQuAD v1.1}} \\ 
		\cmidrule{3-5}\cmidrule{6-8}
		&  & ROUGE-1$\uparrow$  & ROUGE-2$\uparrow$ & ROUGE-L$\uparrow$ & ROUGE-L$\uparrow$  & BLEU-4$\uparrow$ & METEOR$\uparrow$ \\ 
		\midrule
		\multirow{4}{*}{\textbf{AR}} & \textbf{Transformer} & 30.66 & 10.80 & 24.48 & 29.43 & 4.61 & 9.86 \\
		& \textbf{MASS} & 39.70 & 17.24 & 31.91 & 49.48 & 20.16 & 24.41 \\
		& \textbf{ProphetNet} & 39.89 & 17.12 & 32.07 & 48.00 & 19.58 & 23.94 \\ 
		& \textbf{BART} & 38.79 & 16.16 & 30.61 & 42.55 & 17.08 & 23.19  \\ 
		\cmidrule{1-8}
		\multirow{5}{*}{\textbf{Semi-NAR}} & \textbf{InsT} & 17.65 & 5.18 & 16.05 & 29.98 & 2.34 & 8.15  \\
		& \textbf{iNAT} & 26.95 & 6.88 & 22.43 & 32.34 & 3.16 & 9.18  \\
		& \textbf{LevT} & 25.33 & 7.40 & 21.48 & 30.81 & 2.68 & 9.40  \\
		& \textbf{CMLM} & 29.12 & 7.70 & 23.04 & 29.60 & 3.89 & 9.70  \\
		& \textbf{BANG} & 34.71 & 11.71 & 29.16  & \textbf{47.39} & \textbf{17.62} & \underline{21.69}  \\
		\cmidrule{1-8}
		\multirow{7.5}{*}{\textbf{NAR}} & \textbf{NAT} & 24.04 & 3.88 & 20.32 & 31.51 & 2.46 & 8.86 \\
		& \textbf{iNAT} & 24.02 & 3.99 & 20.36 & 32.44 & 2.33 & 8.84  \\
		& \textbf{CMLM} & 23.82 & 3.60 & 20.15 & 31.58 & 2.51 & 8.85  \\
		& \textbf{LevT} & 24.75 & 4.18 & 20.87 & 31.38 & 2.27 & 9.14  \\
		& \textbf{BANG} & 32.59 & 8.98 & 27.41 & 44.07 & 12.75 & 18.99  \\
		& \textbf{ELMER} & \underline{38.30} & \underline{14.17} & \underline{29.92} & 40.22 & 13.49 & 20.08 \\
  \cmidrule{1-8}
  \textbf{Diffusion} & \textbf{Ours} & \textbf{38.84} & \textbf{15.30} & \textbf{30.88} & \underline{46.64} & \underline{16.19} & \textbf{21.99} \\
		\bottomrule   
	\end{tabular}
	\caption{The comparison between different methods on XSUM and SQuAD v1.1 datasets. \textbf{Bold} and \underline{underline} fonts denote the best and second best methods within NAR and Semi-NAR models, respectively. The baseline results are collected from \cite{DBLP:conf/icml/QiG0YCLTLCZ0D21} and \cite{DBLP:journals/corr/abs-2210-13304}.}
	\label{tab:xsum-squad}
\end{table*}

\begin{table*}[t]
	\small
	\centering
	\begin{tabular}{l r r r r r r r}
		\toprule
		\multirow{2.5}{*}{\textbf{Models}} & \multicolumn{3}{c}{\textbf{MSNews}} & \multicolumn{3}{c}{\textbf{MSQG}} & \textbf{CoQA}\\ 
		\cmidrule{2-8}
		 & ROUGE-1$\uparrow$  & ROUGE-2$\uparrow$ & ROUGE-L$\uparrow$ & ROUGE-L$\uparrow$  & BLEU-4$\uparrow$ & METEOR$\uparrow$ & F1$\uparrow$\\ 
		\midrule
        \textbf{LSTM} & 30.0 & 14.6 & 27.7 & 25.3 & 3.5 & 14.1 & 15.1 \\
		\textbf{Transformer} & 33.0 & 15.4 & 30.0 & 29.3 & 5.1 & 16.6 & 15.7 \\
		\textbf{BART}  &41.8 &23.1 &38.3 & 38.1 & 10.2 & 22.1 &64.6 \\
		\midrule
		\textbf{BANG} &32.7 &\underline{16.1} & 30.3& \underline{33.1} & \textbf{11.0} & \underline{18.4} & 31.4 \\
		\textbf{ELMER} &\underline{35.6} &\underline{16.1} &\underline{32.5} & 26.6 & 5.00 & 15.7 & \underline{63.1} \\
  \midrule
        \textbf{Ours} & \textbf{46.8} & \textbf{31.6} & \textbf{44.2} & \textbf{33.3} & \underline{6.6} & \textbf{19.3} & \textbf{65.4} \\
		\bottomrule   
	\end{tabular}
	\caption{The comparison between different methods on MSNews, MSQG and CoQA datasets. \textbf{Bold} and \underline{underline} fonts denote the best and second best methods within NAR models, respectively.}
	\label{tab:msn}
\end{table*}

\subsection{Experimental Results}
We report the main experimental results of our Diffusion-NAT and baselines on seven text generation tasks in Table~\ref{tab:personachat}, Table~\ref{tab:xsum-squad} and Table~\ref{tab:msn}.

\paragraph{Dialog Generation.}
As shown in Table~\ref{tab:personachat}, for the coherence metrics (\ie BLEU-1/2), the performance order of aforementioned baselines in the two dialog generation datasets is mostly consistently as: \textit{AR models} > \textit{Semi-NAR models} > \textit{NAR models}.
It indicates that AR models are more capable of generating coherent and fluent responses than NAR ones.
A major reason is that AR models can better capture the dependency of tokens.
Whereas, for the diversity metrics, AR models mostly underperform NAR models.
The reason may be that AR models are easy to overfit into the frequently co-occurring tokens (\eg I am OK.) in the training data, causing the ``safe response'' problem.
Besides, we observe that the NAR methods using pre-training technique (\ie BANG and ELMER) can better balance the coherence and diversity metrics, and greatly outperform other NAR models.
It demonstrates the effectiveness of large-scale pre-training in improving the NAR generation performance.

Finally, Diffusion-NAT mostly outperforms Semi-NAR and NAR models on all metrics.
Different from these baselines, our approach is based on the discrete diffusion model that can iteratively refine the generated results using a PLM BART.
As we have adapted them to better fit with each other by a set of revisions, we can combine the merits of the rich knowledge from BART and the iterative refining mechanism of the diffusion model.
In this way, we can improve both the coherence and diversity of the generated responses.
Furthermore, our approach outperforms AR models in the overall metric, \eg Ours (27.90) VS. BART (23.54) in PersonaChat.
The reason is that our approach achieves much higher values in the Distinct-1,2 metrics. It shows the effectiveness of our approach for generating diverse responses.


\paragraph{Text Summarization and Question Generation.}
As shown in Table~\ref{tab:xsum-squad} and Table~\ref{tab:msn}, AR models outperform NAR models in a large margin.
The reason is that the two types of tasks mainly require the model to accurately generate proper texts, which is more suitable for AR models due to their superiority of capturing the token dependency.
Despite this, our approach mostly outperforms all the NAR and Semi-NAR methods, and even surpasses AR models on part of datasets (\eg MSNews). 
It is because our approach can effectively combine the merits of the PLM that has pre-learned rich semantic knowledge and the diffusion models that can iteratively refine the produced results, leading to higher-quality generated texts.

\paragraph{Conversational Question Answering.}
The conversational question answering task is to evaluate both the generative capacity and the world knowledge of the model.
As shown in Table~\ref{tab:msn}, our approach also performs well in this task, even slightly outperforms the AR model BART by 0.8 on F1 metric.
A possible reason is that our approach can make use of the pre-learned world knowledge from BART. Besides, as our model can also leverage the iterative refining paradigm of the diffusion model, it may also fix the wrong answers in the generated text, leading to more accurate answers.

\begin{table}[t]
	\small
	\centering
	\begin{tabular}{lccc}
		\toprule
		 \multirow{2.5}*{\textbf{Models}} & \multicolumn{3}{c}{\textbf{PersonaChat}} \\
		 \cmidrule{2-4}
		 & \textbf{Fluency} & \textbf{Informativeness} & \textbf{Relevance} \\
		\midrule
		BART & 4.32 & 4.31 & 3.47  \\
		ELMER & 3.88 & 4.49 & 2.90  \\
		Ours & 4.29 & 4.57 & 3.19      \\
		\bottomrule
	\end{tabular}
	\caption{Human evaluation scores of different methods about the generated responses on PersonaChat.}
	\label{tab:turing-test}
\end{table}

\paragraph{Human Evaluation.}
In addition to the automatic metrics, human evaluation is also critical for text generation.
Considering the expensive annotation cost, we only focus on the dialog generation task and compare our approach with two best performing baselines, \ie BART and ELMER.
Following existing works~\cite{DBLP:journals/corr/abs-2210-13304}, we randomly select 500 examples from the test set of the PersonaChat dataset, and invite three annotators to evaluate the quality of the generated responses from the two baseline and ours from the perspectives of Fluency, Informativeness and Relevance.
The scoring range is from 1 to 5.

As shown in Table~\ref{tab:turing-test}, we can see that the AR method BART performs better on the Fluency and Relevance metrics while the NAR method ELMER performs well on informativeness.
Such results show a similar tendency as the automatic metrics, and indicate the different superiority of AR and NAR models.
As a comparison, our approach can well balance the three metrics, with the comparable performance on Fluency as BART and the best performance on Informativeness.
It shows the great potentiality of discrete diffusion models with PLMs in NAR text-to-text generation tasks.

\section{Conclusion}
In this paper, we proposed Diffusion-NAT, a self-prompting discrete diffusion model~(DDM) using a PLM BART for non-autoregressive~(NAR) text generation.
In our approach, we unified the inference process of BART and the denoising process of DDM into the same masked tokens recovering task, to combine the merits of both the rich pre-learned knowledge of BART and the iterative refining paradigm of DDM.
Concretely, we revised the decoding process of BART into the NAR manner, and adapted the typical settings of DDM to better fit with BART, including Markov transition matrix, training objective and time step embeddings.
Besides, we devised an iterative self-prompting strategy to guide the PLM to deliberate and refine the intermediate generated results, to further improve the quality of final produced texts.
Extensive experiments on seven datasets have shown that our approach can outperform competitive NAR and Semi-NAR models, and even surpass AR models.

In future work, we will investigate more effective and efficient way to combine PLMs and DDM for NAR text generation, \eg prompt learning.

\section*{Limitations}
This work is to investigate discrete diffusion models with pre-trained language models for non-autoregressive text-to-text generation.
An important limitation is the relatively higher inference latency of diffusion models.
In this work, we have adopted DDIM to accelerate the inference process by reducing the diffusion steps, and we also conduct experiments to investigate the performance changes w.r.t. different steps in Appendix~\ref{app-hyper}.
We can see that fewer steps using DDIM would lead to the performance degradation.
Fortunately, there are several recent works that have shown effectiveness in solving this problem~\cite{lu2022dpm}.
As these methods are general to all diffusion models, they may be able to be utilized in our approach.
Besides, as we have adopted a PLM, BART in our approach, it may present biases learned from the pre-training corpus in the generated texts.

\bibliography{custom}
\bibliographystyle{acl_natbib}
\newpage
\appendix

\section{Details of Datasets}
\label{app-dataset}
We conduct experiments on seven datasets, corresponding to four representative text generation tasks. The statistics of these datasets are shown in table~\ref{tab:statistics}. 
\begin{itemize}
    \item \textbf{Dialog Generation}
    aims to predict responses according to the dialog history. We select \textbf{DailyDialog} \cite{DBLP:conf/ijcnlp/LiSSLCN17} and \textbf{PersonaChat} \cite{DBLP:conf/acl/KielaWZDUS18} datasets.
    \item \textbf{Text Summarization} is to summarize the document into a sentence. We choose \textbf{XSUM} \cite{DBLP:conf/emnlp/NarayanCL18} and  \textbf{MSNews} \cite{DBLP:conf/acl/LiuYGQZJCFSGWCJ21}, two news summarization datasets.
    \item \textbf{Question Generation} aims to generate questions based on given passages and answers. We use \textbf{MSQG} \cite{DBLP:conf/acl/LiuYGQZJCFSGWCJ21} and \textbf{SQUAD v1.1} \cite{DBLP:conf/emnlp/RajpurkarZLL16} datasets.
    \item \textbf{Conversational Question Answering} is to answer the question based on a conversation. We select \textbf{CoQA} \cite{DBLP:journals/tacl/ReddyCM19} dataset.
\end{itemize}

\section{Details of Baselines}
\label{app-baselin}
We mainly compare our Diffusion-NAT with a variety of Semi-NAR and NAR models.
\textbf{NAT} \cite{DBLP:conf/iclr/Gu0XLS18}, \textbf{iNAT} \cite{DBLP:conf/emnlp/LeeMC18}, \textbf{InsT} \cite{DBLP:conf/icml/SternCKU19}, \textbf{CMLM} \cite{DBLP:conf/emnlp/GhazvininejadLL19} and \textbf{LevT} \cite{DBLP:conf/nips/GuWZ19} are five Transformer-based NAR models with special generation strategies, \ie iterative refinement, conditional masked language modeling and insertion-deletion operation. 
\textbf{BANG} \cite{DBLP:conf/icml/QiG0YCLTLCZ0D21} and \textbf{ELMER} \cite{DBLP:journals/corr/abs-2210-13304} adopt the pre-training technique based on Transformer to further improve the NAR generation performance.
Note that InsT, iNAT, LevT, CMLM and BANG also support the semi-NAR manner that can rely on partially generated results for improving the inference.

We also compare our approach with AR models which have shown better accuracy than NAR ones.
\textbf{LSTM} \cite{DBLP:journals/neco/HochreiterS97} and \textbf{Transformer} \cite{DBLP:conf/nips/VaswaniSPUJGKP17} are two classic Seq2Seq models. \textbf{MASS} \cite{DBLP:conf/icml/SongTQLL19}, \textbf{BART} \cite{DBLP:conf/acl/LewisLGGMLSZ20} and \textbf{ProphetNet} \cite{DBLP:conf/emnlp/QiYGLDCZ020} are PLMs specially for text generation and we use their base version for fair comparison.

\section{Implementation Details}
\label{app-details}
For all baselines, we use the source code provided by their authors, and all hyper-parameters are set following the suggestions from the original paper.
For our Diffusion-NAT, we use the checkpoint of BART-base with 110M parameters for initialization, and do not add any other parameters.
We use the linear noise schedule~\cite{DBLP:conf/nips/HoJA20} for the diffusion process.
During training, the diffusion step is set to 1000.
During inference, we utilize DDIM~\cite{DBLP:conf/iclr/SongME21} for fast sampling and reduce the diffusion step into 100.
The number of self-prompting turns is set to 2.
We use AdamW as the optimizer, and set learning rate to 5e-5. We set the training step for XSUM and SQuAD v1.1 to 120k, and 80k for other datasets.
The batch size is set to 512.
All experiments are conducted on 8 NVIDIA Tesla V100 GPUs.
The training process of each task requires less than 24 hours.

\begin{table}[t]
	\small
	\centering
	\setlength\tabcolsep{3.5pt}
	\begin{tabular}{lrrrr}
		\toprule
		 \multirow{2.5}*{\textbf{Models}} & \multicolumn{4}{c}{\textbf{PersonaChat}} \\
		 \cmidrule{2-5}
		 & \textbf{B-1} & \textbf{B-2} & \textbf{D-1} & \textbf{D-2} \\
		\midrule
		ELMER    & 31.11 & 23.99 & 3.66 & 24.96 \\
		\midrule
		Ours       & 44.55 & 37.66 & 3.19 & 26.20 \\
		\midrule
		-w/o self-prompting & 43.93 & 37.19 & 2.62 & 22.22 \\
        -w/o PLM & 41.39 & 35.33 & 1.74 & 17.31 \\
        +Time step Embedding & 40.03 & 33.80 & 1.75 & 16.80 \\
        BART=>RoBERTa & 38.07 & 32.17 & 2.99 & 18.32 \\
		\bottomrule
	\end{tabular}
	\caption{Ablation study on PersonaChat dataset.}
	\label{tab:ablation}
\end{table}

\begin{table}[t]
	\small
	\centering
	\setlength\tabcolsep{3.5pt}
	\begin{tabular}{lccccc}
		\toprule
		 \multirow{2.5}*{\textbf{Models}} & \multicolumn{2}{c}{\textbf{PersonaChat}} & \textbf{XSUM} & \multicolumn{2}{c}{\textbf{SQuAD}} \\
		 \cmidrule{2-6}
		 & \textbf{B-1} & \textbf{B-2} & \textbf{R-L} & \textbf{R-L} & \textbf{MT} \\
		\midrule
		DiffuSeq    & 37.79 & 32.50 & 20.29 & 29.29 & 12.57 \\
		\midrule
		Ours       & 44.55 & 37.66 & 30.88 & 46.64 & 21.99 \\
		\bottomrule
	\end{tabular}
	\caption{Performance comparison of continuous diffusion method DiffuSeq~\cite{DBLP:journals/corr/abs-2210-08933} and our approach on PersonaChat, XSUM and SQuAD datasets.}
	\label{tab:vs}
\end{table}

\section{Ablation and Variation Study}
Our Diffusion-NAT includes several key designs, \ie the usage of BART, self-prompting strategy, removing time step embeddings.
Here, we conduct the ablation and variation study on our approach to verify their effectiveness.
Concretely, we propose four variations of our approach as shown in Table~\ref{tab:ablation}, where \textbf{-w/o self-prompting} and \textbf{-w/o PLM} refer to the variations removing the corresponding component, \textbf{+Time step Embeddings} and \textbf{BART=>RoBERTa} are the variations that add the time step embeddings as continuous diffusion methods~\cite{DBLP:journals/corr/abs-2105-10311} and replaces BART by RoBERTa in our approach, respectively.
We can see that all the variations underperform our approach, it demonstrates the effectiveness of the above designs.
Among them, we can see that adding time step embeddings cause the performance degrading a lot. The reason is that the additional embeddings may disturb the original semantic representations of BART.

\section{Discrete Diffusion V.S. Continuous Diffusion}
For the NAR text-to-text generation, existing works~\cite{DBLP:journals/corr/abs-2210-08933} also have incorporated the continuous diffusion method.
In this part, we aim to compare our approach with a recently proposed work, DiffuSeq~\cite{DBLP:journals/corr/abs-2210-08933} that performs continuous diffusion on the latent space of token embeddings and leverages the KNN rounding step to map the embeddings into discrete tokens.
We conduct the experiments on PersonaChat, XSUM and SQuAD datasets.
As shown in Table~\ref{tab:vs}, we can see that our approach outperforms DiffuSeq in all metrics by a large margin.
It shows the effectiveness of our proposed method that utilizes the discrete diffusion method in NAR text-to-text generation tasks.
Besides, compared with DiffuSeq, our approach can also benefit from the PLM BART, which also helps generate higher-quality texts.

\begin{figure}[t]
	\centering
	\subfigure[BLEU-2]{\label{fig-csqa-control}
		\centering
		\includegraphics[width=0.225\textwidth]{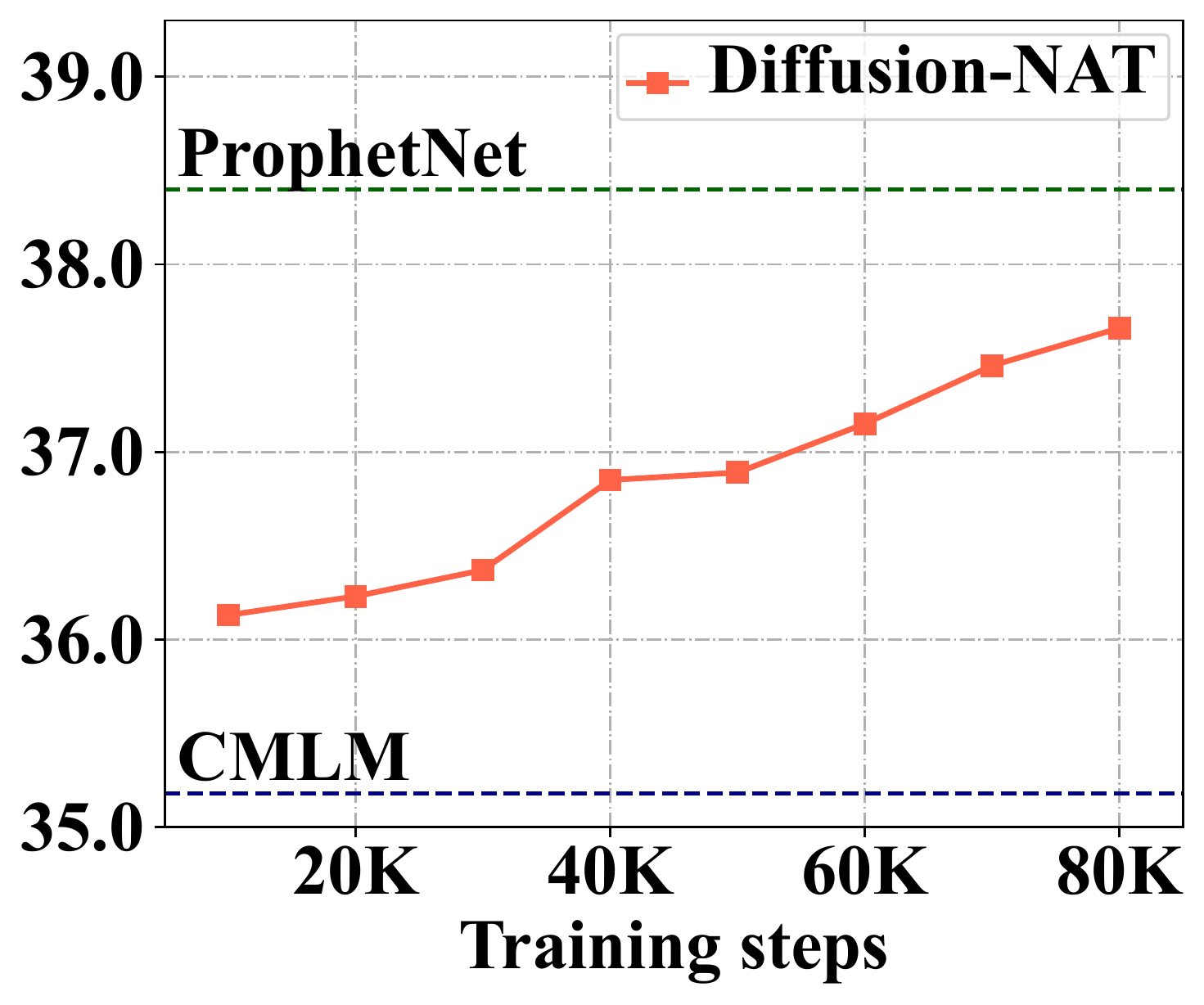}
	}
	\subfigure[Distinct-2]{\label{fig-obqa-control}
		\centering
		\includegraphics[width=0.225\textwidth]{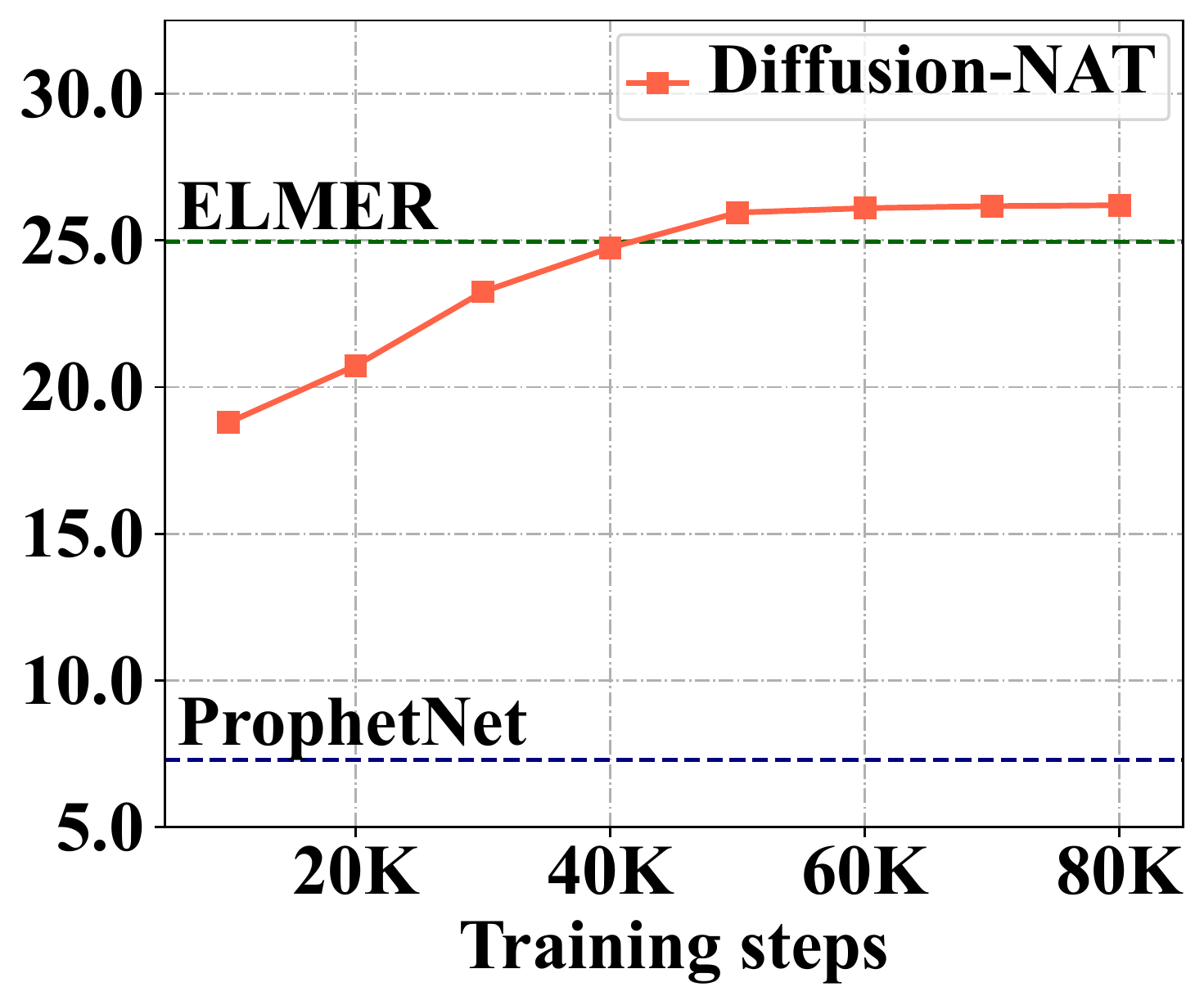}
	}
	\centering
	\caption{Performance changes of our approach w.r.t. the training steps on PersonaChat dataset.}
	\label{fig-training}
\end{figure}

\section{Performance w.r.t. Training Steps}
As our approach adopts the pre-trained BART for parameters initialization, it is also helpful to faster and better convergence.
To verify it, we report the BLEU-2 and Distinct-2 performance changes of our approach w.r.t. the training steps during training.
As show in Figure~\ref{fig-training}, we observe that with the increasing of training steps, the performance of our approach is consistently improving, gradually approaching or surpassing competitive models.
It shows the stabilization of our convergence process.
Besides, for BLEU-2, with just 10k training steps, our approach can outperform competitive Semi-NAR model CMLM.
The reason may be that BART provides a good starting point of the training process, making our approach converge faster.

\begin{table}[t]
	\small
	\centering
	\setlength\tabcolsep{3.5pt}
	\begin{tabular}{lcccccc}
		\toprule
		  & \multicolumn{6}{c}{\textbf{PersonaChat}}  \\
    \midrule
		 Diff. Steps & 2 & 10 & 20 & 100 & 200 & 1000 \\
		\midrule
		BLEU-2 & 30.82 & 35.88 & 36.19 & 37.66 & 37.63 & 37.65 \\
		\midrule
		Distinct-2 & 23.68 & 27.54 & 26.93 & 26.20 & 26.35 & 26.39 \\
		\bottomrule
	\end{tabular}
	\caption{Performance changes w.r.t. the diffusion steps (abbreviated as Diff. Steps) on PersonaChat dataset.}
	\label{tab:dif-turns}
\end{table}

\begin{table}[t]
	\small
	\centering
	\setlength\tabcolsep{3.5pt}
	\begin{tabular}{lcccccc}
		\toprule
		  & \multicolumn{6}{c}{\textbf{PersonaChat}}  \\
    \midrule
		 SP Turns & 0 & 1 & 2 & 3 & 4 & 5 \\
		\midrule
		BLEU-2    & 35.00 & 36.50 & 37.66 & 37.69 & 37.77 & 37.77 \\
		\midrule
		Distinct-2  & 26.01 & 26.22 & 26.20 & 26.34 & 26.29 & 26.30 \\
		\bottomrule
	\end{tabular}
	\caption{Performance changes w.r.t. the self-prompting turns (abbreviated as SP Turns) on PersonaChat dataset.}
	\label{tab:sp-turns}
\end{table}

\section{Hyper-parameter Tuning.}
\label{app-hyper}
Our approach also requires some parameters to tune, \ie the diffusion steps during decoding and the turns of self-prompting.
Generally, more diffusion steps and self-prompting turns would lead to better performance but larger inference latency, hence we can tune their values to balance the inference time cost and quality.
In this part, we conduct experiments on the PersonaChat dataset to validate it.
As shown in Table~\ref{tab:dif-turns} and Table~\ref{tab:sp-turns}, we can see that more diffusion steps and more self-prompting turns are able to improve the model performance, while the improvement seems to be saturated after a certain number, \ie 100 for diffusion steps and 2 for self-prompting turns.
Such results can provide a reference for tuning the two hyper-parameters to match the requirement of model performance and inference latency.
Besides, we can see that with very few diffusion steps (\eg 2 steps), our approach can also achieve a decent performance on BLEU-2 and Distinct-2.
It demonstrates the potentiality of further reducing the inference latency in our approach.

\begin{table*}[t]
\centering
\small
\begin{tabular}{l|p{5.5cm}|p{5.5cm}}
\toprule
\hline
Personal Profile & i enjoy cross stitch and quilting. my father served in our military in the war. i am proud to be an american. i am what people call a baby boomer. my parents were proud immigrants to america. & my father was a police officer. my favorite band is nirvana. i like running. i work at pet smart. i am a musician. \\ \midrule
Context  & hello, how is your day going hi. my day is good. i am hanging out with both of my sons. cool, sounds like fun. just as much fun as cross stitch and quilts i love those crafts! right now i am into my aquarium filled with exotic fish. i am what you would call a baby boomer, explains the quilting passion you might like my collection of decorations. they are 50s vintage! sure, it would bring back old memories. especially when my dad served in the military that is awesome. what do you eat for breakfast? mine is granola every day. oatmeal fan over here. my parents were immigrants, they raised me on oats everyday. that is very healthy. i like it. do you eat fish? & hey. want to chat? i am just listening to my favorite music, nirvana. i was just reading a biography. i love those. i work too much overtime at pet smart to read. what do you do? i just got out of college. \\ \midrule
Real Response  & yes i do eat fish. i love it & what did you study. i work at pet smart but really like music.\\ \midrule
Ours &fish is almost as healthy as american lifestyle, love fish too & i am a musician. and i play music all the time.\\ \midrule  
\bottomrule
\end{tabular}	
    \caption{Examples of generated responses on PersonaChat by our approach.}	
    \label{tab-case}
\end{table*}

\section{Case Study}
To provide the qualitative analysis on our approach, we show two generated examples on the PersonaChat dataset in Table~\ref{tab-case}.
We can see that with the help of BART and the diffusion model, our approach can generate relevant and informative responses based on the given dialog context.
Besides, the left example shows that our approach is able to generate interesting phrases such as ``as healthy as american lifestyle'', which makes the response more humorous and also well reflects the speaker's personal characteristics.

\end{document}